\documentclass[10pt,twocolumn,letterpaper]{article}

\usepackage{cvpr}              

\usepackage{times}
\usepackage{graphicx}
\usepackage{amsmath,amssymb,amsthm,mathabx}
\usepackage{booktabs}
\usepackage{algorithmic}
\usepackage[linesnumbered,ruled,vlined]{algorithm2e}
\usepackage{acronym}
\usepackage{enumitem}
\usepackage[pagebackref,breaklinks,colorlinks]{hyperref}
\usepackage{balance}
\usepackage{xspace}
\usepackage{setspace}
\usepackage[dvipsnames]{xcolor}
\usepackage[capitalise]{cleveref}
\usepackage{tabularx,colortbl,multirow,array,makecell}
\usepackage{cuted}
\usepackage{capt-of}

\makeatletter
\DeclareRobustCommand\onedot{\futurelet\@let@token\@onedot}
\def\@onedot{\ifx\@let@token.\else.\null\fi\xspace}
\def\eg{\emph{e.g}\onedot} 
\def\Eg{\emph{E.g}\onedot}
\def\ie{\emph{i.e}\onedot}

\def\etc{\emph{etc}\onedot} 
\def\vs{\emph{vs}\onedot}

\def\etal{\emph{et al}\onedot}
\makeatother

\makeatletter
\renewcommand{\paragraph}{%
  \@startsection{paragraph}{4}%
  {\z@}{0ex \@plus 0ex \@minus 0ex}{-1em}%
  {\hskip\parindent\normalfont\normalsize\bfseries}%
}
\makeatother

\graphicspath{{figures/}}

\crefname{algocf}{Alg.}{Algs.}
\Crefname{algocf}{Algorithm}{Algorithms}
\crefname{section}{Sec.}{Secs.}
\Crefname{section}{Section}{Sections}
\crefname{table}{Tab.}{Tabs.}
\Crefname{table}{Table}{Tables}

\definecolor{gblue}{HTML}{4285F4}
\definecolor{gred}{HTML}{DB4437}
\definecolor{ggreen}{HTML}{0F9D58}

\acrodef{ai}[AI]{Artificial Intelligence}

\usepackage{amsmath,amsfonts,bm}

\def\eqref#1{equation~\ref{#1}}

\def\1{\bm{1}}

\def\vp{{\bm{p}}}

\def\vr{{\bm{r}}}
\def\vs{{\bm{s}}}

\DeclareMathAlphabet{\mathsfit}{\encodingdefault}{\sfdefault}{m}{sl}
\SetMathAlphabet{\mathsfit}{bold}{\encodingdefault}{\sfdefault}{bx}{n}

\newcommand{\R}{\mathbb{R}}

\def\cvprPaperID{2767} 
\def\confName{CVPR}
\def\confYear{2022}

\begin{document}
\graphicspath{{figures/}}


\title{PartAfford: Part-level Affordance Discovery from 3D Objects}

\author{
Chao Xu\textsuperscript{1} \quad 
Yixin Chen\textsuperscript{1} \quad 
He Wang\textsuperscript{2,4} \quad 
Song-Chun Zhu\textsuperscript{2,3,4} \quad 
Yixin Zhu\textsuperscript{2,4} \quad 
Siyuan Huang\textsuperscript{4} \quad \\
\textsuperscript{1} University of California, Los Angeles\quad
\textsuperscript{2} Peking University \\
\textsuperscript{3} Tsinghua University \quad 
\textsuperscript{4} Beijing Institute for General Artificial Intelligence
}

\twocolumn[{%
\renewcommand\twocolumn[1][]{#1}%
\maketitle
\begin{center}
    \centering
    \vspace{-18pt} 
    \captionsetup{type=figure}
    \includegraphics[width=\textwidth]{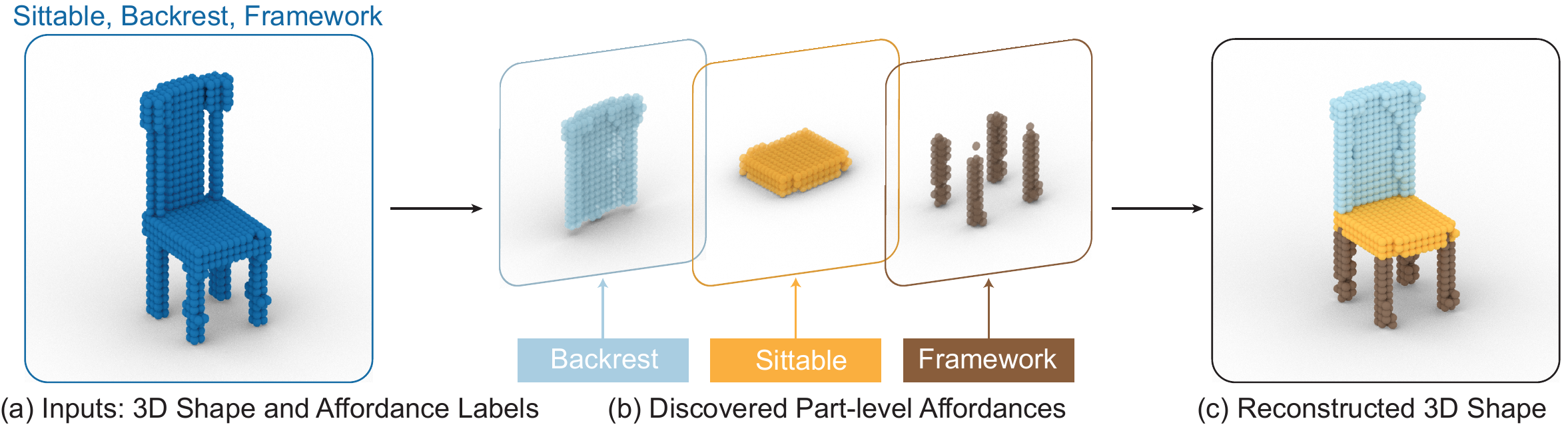}
    \captionof{figure}{\textbf{The proposed \textit{PartAfford} task.} (a) Given 3D shape and its affordance labels as the inputs, we devise algorithms to (b) decompose the 3D object into parts and (c) discover the part-level affordances.}
    \label{fig:illustration}
\end{center}%
}]

\setstretch{0.96}

\begin{abstract}
\vspace{-12pt}
Understanding what objects could furnish for humans--namely, learning object \textit{affordance}--is the crux to bridge perception and action. In the vision community, prior work primarily focuses on learning object affordance with dense (\eg, at a per-pixel level) supervision. In stark contrast, we humans learn the object affordance \emph{without} dense labels. As such, the fundamental question to devise a computational model is: What is the natural way to learn the object affordance from visual appearance and geometry with humanlike sparse supervision? In this work, we present a new task of \textbf{part-level affordance discovery (PartAfford)}: Given only the affordance labels per object, the machine is tasked to (i) decompose 3D shapes into parts and (ii) discover how each part of the object corresponds to a certain affordance category. We propose a novel learning framework for \emph{PartAfford}, which discovers part-level representations by leveraging only the affordance set supervision and geometric primitive regularization, without dense supervision. The proposed approach consists of two main components: (i) an abstraction encoder with slot attention for unsupervised clustering and abstraction, and (ii) an affordance decoder with branches for part reconstruction, affordance prediction, and cuboidal primitive regularization. To learn and evaluate PartAfford, we construct a part-level, cross-category 3D object affordance dataset, annotated with $24$ affordance categories shared among $>25,000$ objects. We demonstrate that our method enables both the abstraction of 3D objects and part-level affordance discovery, with generalizability to difficult and cross-category examples. Further ablations reveal the contribution of each component.

\end{abstract}

\setstretch{1}

\section{Introduction}

The human vision system could swiftly locate the functional part upon using an object for specific tasks~\cite{land1999roles}. Such a critical capability in object interaction requires fine-grained object \textit{affordance} understanding. \textit{Affordance}, coined and originally theorized by Gibson~\cite{gibson1966senses,gibson1979ecological}, characterizes how humans interact with human-made objects and environments. As such, affordance understanding of objects and scenes has a significant influence on bridging visual perception and holistic scene understanding~\cite{huang2018holistic,huang2018cooperative,chen2019holistic++} with actionable information~\cite{soatto2013actionable}. It is considered as one of the critical ingredients for the artificial general intelligence~\cite{zhu2020dark}.

Object affordances have two main characteristics. First, object affordances are not defined in terms of conventional categorical labels in computer vision; instead, they are defined by the associated actions for various tasks and are naturally \textit{cross-category}. For example, both chair and sofa can be sat on, which indicates they share the \textit{sittable} affordance. Similarly, desktop and bookshelf share the \textit{support} affordance. Second, object affordances are intrinsically \textit{part-based}. We could easily associate \textit{sittable} affordance with the seats of chairs and sofas, and \textit{support} with the boards of desktop and bookshelf. As such, the ability to learn \textbf{part-based}, \textbf{cross-category} affordance is essential to demonstrate the general object affordance understanding.

In passive affordance learning, prior literature follows the supervised learning paradigm, in which dense affordance annotation on the objects is fed as supervised signals~\cite{deng20213d}. However, this line of thought depends heavily on the quality of dense annotation, significantly deviated from how we humans learn to understand affordance. A humanlike supervision would be: ``you can sit on this chair and rest your arm,'' ``you can open the lid and hold water with the cup.'' In this paper, we try to answer: How to distinguish each object part while recognizing corresponding affordances with such sparse and natural supervisions?

To tackle this problem, we present \textit{PartAfford}, a new task of part-level affordance discovery, which learns the object affordance with natural supervision of affordance set. As shown in \cref{fig:illustration}, by providing only the set of affordance labels for each object, the algorithm is tasked to decompose the 3D shapes into parts and discover how each part corresponds to a certain affordance category.

\textit{PartAfford}'s core challenges lie in the visual ambiguity of 3D part-level segmentation and reconstruction under sparse supervision. To address it, we propose a novel method that discovers part-level representations with affordance set supervision and primitive regularization. The proposed approach consists of two main components. The first component is an encoder with slot attention for unsupervised clustering and abstraction. Specifically, we encode the 3D object into visual features and abstract the low-level features into a set of \textit{slot} variables~\cite{locatello2020object}. The second component is a decoder built upon the learned slot features. It has three output branches that jointly reconstruct the 3D parts and object, predict the affordance labels, and regularize the learned part-level shapes with cuboidal primitives. Our method does not rely on dense supervision but instead learns from the sparse set supervision. It discovers the part-level affordance by learning the correspondence between affordance labels and abstracted 3D objects.

Learning and evaluating \textit{PartAfford} demands collections of 3D objects and their affordance labels for object parts. Prior work on visual affordance learning~\cite{hassanin2021visual} either focuses on 2D objects and scenes or lacks part-based annotation~\cite{deng20213d}. Hence, we construct a part-level, cross-category 3D object affordance dataset annotated with $24$ affordance categories shared among over $25,000$ 3D objects. The 3D objects are collected from PartNet dataset~\cite{mo2019partnet} and the PartNet-Mobility dataset~\cite{xiang2020sapien}. The $24$ part affordance categories are defined in terms of adjectives (\eg, ``sittable'') or nouns (\eg, ``armrest''); they describe how object parts could afford human daily actions and activities. We annotate the part-level object affordances by manually mapping the fine-grained object part defined in Mo \etal~\cite{mo2019partnet} to the part affordances defined in this work.

By experimenting on this newly constructed \textit{PartAfford} dataset, we empirically demonstrate that our method jointly enables the abstraction of 3D objects and part-level affordance discovery. Our model also shows strong generalizability on hard and cross-category objects. Further experiments and ablations analyze each component's contribution and point out future directions.

\setstretch{0.97}

In summary, our work makes four main contributions:

\begin{itemize}[leftmargin=*,noitemsep,nolistsep]
    \item We present a new \textit{PartAfford} task for part-level affordance discovery. Compared to the prior densely-supervised learning paradigm, this approach learns the visual object affordance in a more natural manner.
    \item We propose a novel method for tackling \textit{PartAfford}, which jointly abstracts 3D objects into part-level representations and discovers the affordance by learning the affordance correspondence.
    \item We construct a dataset consisting of 3D objects and annotate part-level affordances to support the learning and evaluation of \textit{PartAfford}.
    \item We empirically demonstrate the efficacy and generalization capability of the proposed method and analyze each component's significance via a suite of ablation studies.
\end{itemize}

\section{Related Work}

\paragraph{Affordance Learning}

Affordance learning is a multidisciplinary research field of vision, cognition, and robotics. In general, ``affordance'' is first perceived from images~\cite{gupta20113d,zhu2015understanding,roy2016multi} or videos~\cite{xie2013inferring,zhu2016inferring,fang2018demo2vec,nagarajan2019grounded,nagarajan2020ego}, followed by cognitive reasoning~\cite{zhu2015understanding,zhu2020dark}, and eventually serves for task and motion planning in robotics~\cite{nagarajan2020learning,xu2020deep,mandikal2021learning,mo2021o2o}.
Prior work tackles affordance at various scales with different representations. Although affordance has been studied at the scene level~\cite{zhao2013scene, gupta2015indoor, roy2016multi}, object-level~\cite{nguyen2017object}, and associated with generated human poses~\cite{kim2014shape2pose,zhu2015understanding,wang2017binge}, few attempts study affordance as a 3D shape analysis task~\cite{yu2015fill,liang2016inferring,zhu2016inferring,wang2017transferring} since it would normally require large-scale, high-quality 3D data. 
A notable recent work~\cite{deng20213d} benchmarks several supervised affordance estimation tasks on PartNet~\cite{mo2019partnet} with dense affordance keypoint annotations. In comparison, we study affordance in a weakly supervised manner, such that the affordance discovery will be guided by affordance set matching and geometry abstraction. Additionally, to support the learning, we construct a new part affordance dataset, providing fine-grained, part-level 3D affordance annotations. It is tailored to modern sparse supervision setting and affordance compositionality study.

\paragraph{Unsupervised Object-centric Learning}

Object discovery has been studied in an iterative end-to-end fashion~\cite{greff2017neural,van2018relational,burgess2019monet,engelcke2019genesis,greff2019multi,du2021unsupervised}. Recently, Locatello \etal~\cite{locatello2020object} present the slot attention module, an efficient and generic framework for object-centric representation extraction. It is capable of modeling compositional nature in synthetic scenes with multiple simple geometry shapes~\cite{multiobjectdatasets19}. Subsequently, Stelzner \etal~\cite{stelzner2021decomposing} and Yu \etal~\cite{yu2021unsupervised} apply slot attention on unsupervised 3D-aware scene decomposition, integrating NeRF~\cite{mildenhall2020nerf} as object representations. They demonstrate that slot-based bottleneck could perform reasonably on synthetic multi-view RGB datasets with a textureless background. Our work takes one step further to tackle the challenge of \textit{part-level} affordance discovery of 3D objects; part discovery is more complex than object discovery, primarily due to the ambiguity in the object part segmentation without applying additional constraints. As ~\cite{locatello2020object} notices that “crowded” 2D scenes (w/ overlapping objects) are “especially difficult,” dense connections between parts, especially the small parts, in an object makes PartAfford a crowded 3D scenario with extended difficulty. Fortunately, for man-made objects, affordances are attached to objects at the part level. This observation implies the possibility of combining part discovery and affordance learning with minimal supervision. In this work, we integrate part discovery with affordance estimation, hoping that affordance information would help discover object parts sharing similar affordances.

\paragraph{Geometric Primitive Modeling}

Whereas supervised geometric primitive abstraction methods~\cite{mo2019structurenet,yang2020dsm} require dense hierarchical annotations, unsupervised frameworks using cuboid-based~\cite{tulsiani2017learning,sun2019learning}, superquadrics-based~\cite{paschalidou2019superquadrics,paschalidou2020learning}, or other genus-zero-shape~\cite{deng2020cvxnet,paschalidou2021neural} primitives discover structural information naturally embedded in the geometry. Recently, Yang \etal~\cite{yang2021unsupervised} unsupervisedly learn the cuboid-based shape abstraction with shape co-segmentation. Yet, it relies heavily on the ground-truth point normals for accurate abstraction and lacks semantic representation for object understanding. In our affordance discovery framework, we leverage the cuboidal regularization to refine the reconstructed affordance part, which distinguishes densely connected 3D parts, improving the affordance part discovery. 

\begin{figure*}[t!]
	\centering
	\includegraphics[width=\linewidth]{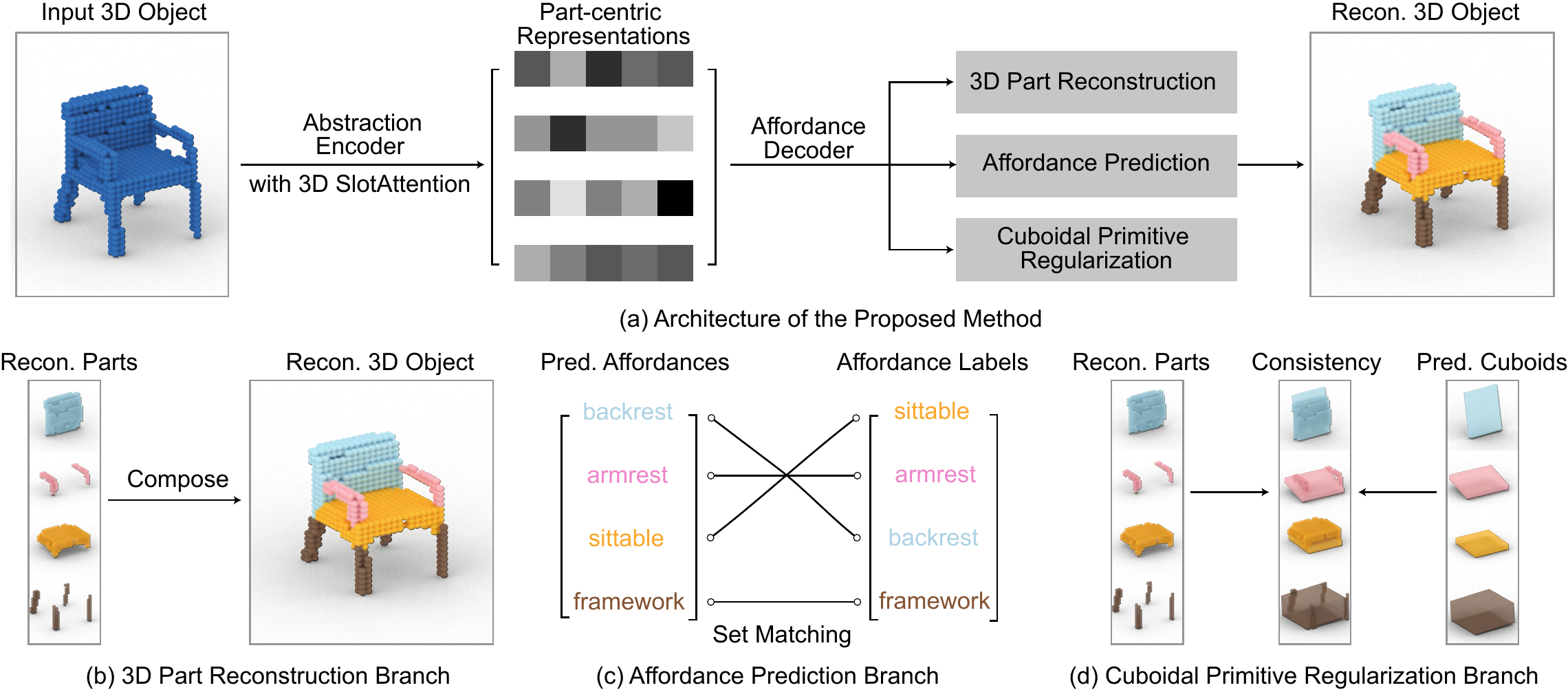}
	\caption{(a) \textbf{Illustration of the proposed method for \textit{PartAfford}}. Our model contains two main components: abstraction encoder and affordance decoder. \textbf{Abstraction encoder} takes 3D voxels as input, extracts features with 3D convolutional neural networks, and abstracts them into several slots. \textbf{Affordance decoder} with three branches jointly (b) reconstructs the 3D parts, (c) predicts affordance labels, and (d) regularizes cuboidal primitives. Given the 3D shape and ground-truth affordance labels, the model can (i) learn the part-level segmentation from the self-supervision of 3D reconstruction, (ii) learn regularized geometric primitives by fitting the reconstructed parts with cuboids, and (iii) discover the corresponding affordance for each slot with set matching loss. Please refer to the \textit{appendix} for more details of the encoder and three branches.}
	\label{fig:decoder}
\end{figure*}

\section{Task Definition}\label{sec:task}

We formulate the new task \textit{PartAfford} as discovering the part-level object affordance with the affordance set supervision. We define $K$ common affordance categories $\mathcal{S} = \{s_k\}_{k=1}^{K}$, such as ``sittable'' and ``openable,'' for indoor object understanding. Input is given as a collection of $N$ objects $\{o_i\}_{i=1}^{N}$ and their corresponding affordance set labels $\{\mathcal{A}_i\}_{i=1}^{N}$, where $\mathcal{A}_i = \{a^j_i\}_{j=1}^{J_i}$. $a_i^j \in \mathcal{S}$ and $J_i$ represents the number of distinct assigned affordances for each object $i$. \textit{PartAfford} requires an algorithm to decompose each object into parts and discover the affordance corresponding to each object part. \cref{fig:illustration} illustrates the \textit{PartAfford} task.

\section{Method}

We propose a novel framework for affordance discovery from 3D objects. It integrates unsupervised part discovery with affordance set prediction and geometric primitive abstraction; see \cref{fig:decoder}. Given a 3D shape represented by voxel grids $\mathcal V$ of resolution $32^3$, our method first encodes the 3D shape into visual features and abstracts it into $M$ slots; each slot represents an abstracted high-level feature for downstream tasks. Next, we utilize a decoder with three branches to jointly (i) decode the features into parts, (ii) predict the affordance label, and (iii) regularize the parts with cuboidal primitives. By composing the 3D parts from the slots with the 3D reconstruction as self-supervision, we ensure the slots combined can depict the entire 3D object. With the set matching loss of affordance prediction, the model discovers the correspondence between parts in slots and the affordance labels. Fitting the reconstructed 3D parts into cuboid representation further regularizes the shape of the part discovery. Below, we describe in detail how each module is constructed and the loss design. 


\setstretch{1}

\subsection{Abstraction Encoder}

The encoder takes 3D voxels as input and abstracts latent codes in an unsupervised manner. It consists of a feature extraction module and a slot attention module~\cite{locatello2020object}; see \cref{fig:decoder}.

\paragraph{Feature Extraction}

The feature embedding backbone encodes the input voxels and generates a $D$ dimensional feature for each voxel. Following common practices~\cite{mescheder2019occupancy}, voxels are encoded by five layers of 3D convolutional neural networks. The embedded feature is then augmented with absolute positional embedding~\cite{locatello2020object}.

\paragraph{3D Slot Attention}

The adopted slot attention architecture~\cite{locatello2020object} serves as the representational bottleneck between the 3D feature embedding network and the downstream decoders. The encoded feature of a 3D shape is fed into an iterative attention module, where $M$ slots are updated for $T$ iterations through a Gated Recurrent Unit~\cite{cho2014learning}. Slot attention is calculated by applying softmax normalization on the dot-product similarity between queries (\ie, linearly-mapped 3D slot features) and keys (\ie, linearly-mapped input features). Thus, those slots compete to attend to part of the input 3D shape according to input geometry. The attention is then applied as the weight for aggregating the values (\ie, linearly-mapped input feature) and updating the slots.

\subsection{Affordance Decoder}

As shown in \cref{fig:decoder}, the affordance decoder takes the slot features as input and consists of three branches for 3D part reconstruction, affordance prediction, and primitive regularization. The parameter of the decoder is shared across slots.

\paragraph{3D Part Reconstruction}

We design a 3-layer 3D transposed convolutional decoder followed by a single MLP layer to reconstruct voxel values $\hat{\mathcal V}^m$ and a voxel mask for each slot. The mask is normalized across slots with softmax, which generates a normalized mask $\hat{\Lambda}^m \in \mathbb{R}^{32\times 32\times 32}$. It is then used to compute the weighted sum of voxel values across slots and combine the reconstructed parts $\{\hat{\mathcal V}^m\}_{m=1}^M$ into a full 3D shape $\hat{\mathcal V}$:
\begin{equation}
    \hat{\mathcal V} = \sum_{m=1}^M \hat{\Lambda}^m \hat{\mathcal V}^m
\end{equation}

The 3D part reconstruction branch is self-supervised by the reconstruction loss between original voxels $\mathcal V$ and reconstructed voxels $\hat{\mathcal V}$; we use the binary cross-entropy (BCE):
\begin{equation}
	\mathcal{L}_{\text{recon}} = \texttt{BCE}(\mathcal V, \hat{\mathcal V})
\end{equation}

\paragraph{Affordance Prediction}

We predict a one-hot affordance label for each slot with a two-layer MLP with sharing weights across slots for classification. 

The affordance prediction branch is weakly-supervised as we do not provide affordance labels for each voxel. Instead, only the affordance label set for the entire object is used as the supervision signal. The model is tasked to learn the alignment between the abstracted parts and the affordance labels from set supervision.

As defined in \cref{sec:task}, the ground-truth set of affordance labels for an input 3D object is denoted as $\mathcal A$. We denote $\hat{\mathcal{A}}$ as the set of slot affordance predictions. $\hat{\mathcal{A}_\sigma}$ is a permutation of elements in $\hat{\mathcal{A}}$, where $\sigma \in \mathfrak G$ and $\mathfrak G$ represent all $M!$ possible permutations. $\mathcal L_{\text{match}}$ is the pairwise matching cost between two sets, which can be calculated by mean square error (MSE) or cross-entropy:
\begin{equation}
	\mathcal{L}_{\text{pred}}  = \min_{\sigma \in \mathfrak G}\mathcal L_{\text{match}}(\mathcal{A}, \hat{\mathcal{A}_{\sigma}}).
	\label{eq:set_loss}
\end{equation}

Due to the order-invariant nature of slot modules, we apply the Hungarian matching algorithm~\cite{kuhn1955hungarian} to calculate the set-based~\cite{carion2020end} affordance prediction loss in \cref{eq:set_loss}.

\paragraph{Cuboidal Primitive Regularization}

As a generalized soft k-means algorithm, the slot attention mechanism heavily relies on visual cues, such as the clustering of pixel colors on the image. As such, it cannot perform precisely in a crowded scene with overlapping objects even on a toy image dataset~\cite{locatello2020object}. In 3D voxel regime, segmenting object into parts is challenging since every voxel is connected to neighboring voxels without distinguishable visual appearances. 

Therefore, we introduce the cuboidal primitive regularization module, providing geometric prior for segmentation: Human-made objects usually have geometric regularity, and cuboid is a concise structural representation for abstraction.

From each slot embedding, the cuboid abstraction module predicts a cuboid parametrized by two vectors~\cite{yang2021unsupervised}: a scale vector $\vs \in \R^3$ and a quaternion vector $\vr \in \R^4$ for 3D rotation. Of note, we calculate the cuboid center from the weighted mean of voxel positions in the slot.

To evaluate how the predicted cuboid fits the reconstructed part in the $m$-th slot, we first compute the Euclidean distance $d_i^m$ from each voxel $\vp^m_i$ to its closest cuboid face.
Next, we calculate the weighted sum distance for all voxels, where the weight $v^m_i \in \hat{\mathcal V}^m$ is the reconstructed voxel value within $[0,1]$. Additionally, we designed a binary surface mask $f(i)$ that masks out internal voxels in the loss. Thus, the loss encourages a cuboid to tightly wrap a solid object. The regularization loss for all the slots is defined as:
\begin{equation}
	\mathcal{L}_{\text{cuboid}} = \sum_m \sum_i f(i) v^m_i d^m_i.
\end{equation}

\subsection{Total Loss}

Taking together, the total training loss is the sum of 3D reconstruction loss, affordance prediction loss, and the primitive regularization loss:
\begin{equation}
	\mathcal{L}_{\text{total}} = \lambda_{\text{recon}}\mathcal{L}_{\text{recon}} + \lambda_{\text{pred}} \mathcal{L}_{\text{pred}} + \lambda_{\text{cuboid}} \mathcal{L}_{\text{cuboid}},
\end{equation}
where $\lambda_{\text{recon}}$, $\lambda_{\text{cuboid}}$, and $\lambda_{\text{pred}}$ are balancing coefficients.

Of note, with the current architecture design, for the first time, we demonstrate the capability of part-level affordance discovery from set labels. The exploration of more complex and practical modules are left for future work.

\setstretch{1}

\section{Part Affordance Dataset}\label{sec:dataset}
\begin{figure}[t!]
	\centering
	\includegraphics[width = \linewidth]{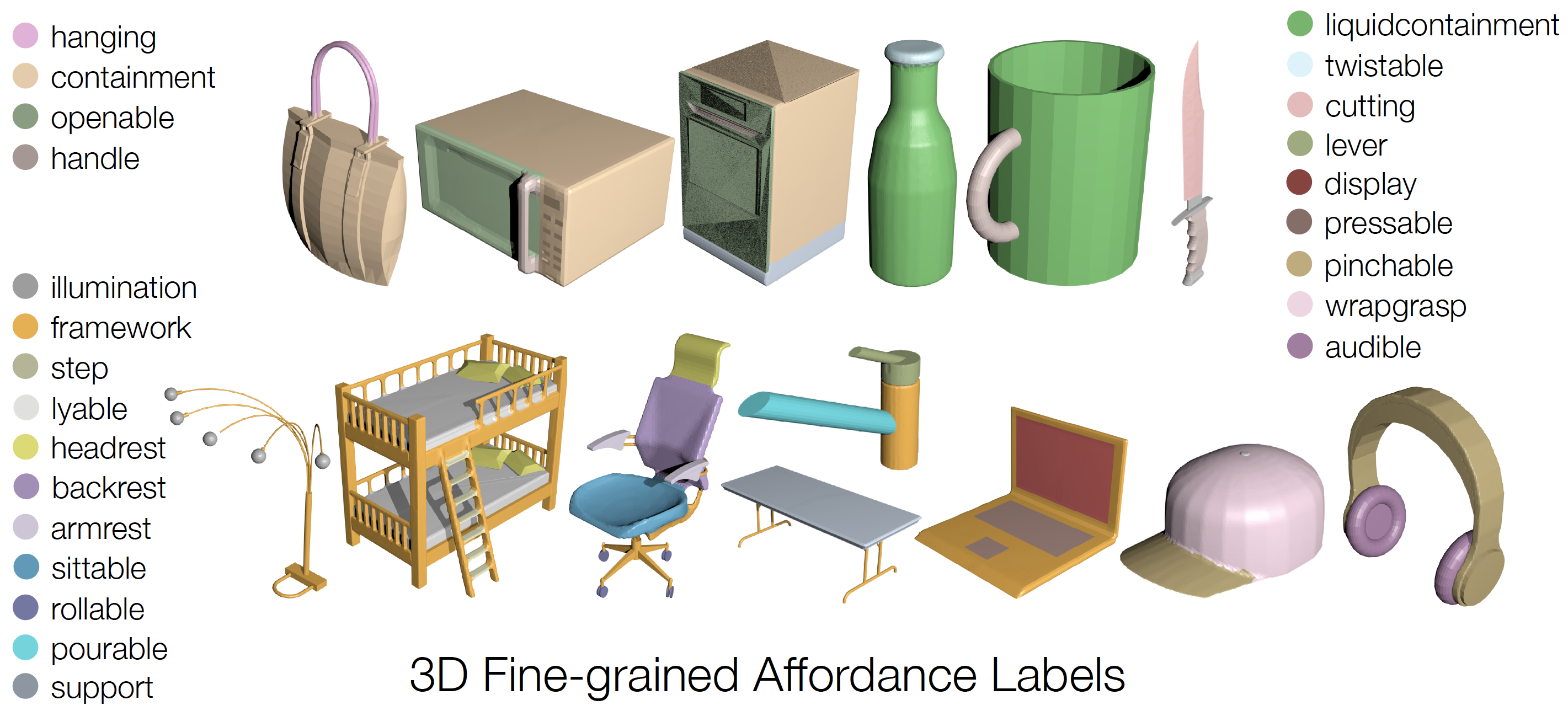}
	\caption{\textbf{Part affordance dataset.} Some exemplar 3D object models are visualized with 24 categories of affordance annotations. Object parts rendered in the same color share the same affordance. Each object part may have multiple affordance labels, and only the affordance label with the highest priority is rendered.}
	\label{affordance_label}
\end{figure}

To facilitate the research in part-level affordance discovery, we construct a part-level 3D object affordance dataset. We focus on $24$ cross-category, fine-grained affordance labels as shown in \cref{affordance_label}. The dataset is annotated with over $25,000$ 3D CAD models from the PartNet dataset~\cite{mo2019partnet} and $625$ articulated objects among 9 categories from the PartNet-Mobility dataset in SAPIEN~\cite{xiang2020sapien}. 
Below, we describe how to define part affordances and the general guideline for affordance annotation. Please refer to the \textit{appendix} for more details.

\subsection{Affordance Definition \& Dataset Construction}

Part affordances in our dataset are defined in terms of adjectives (\eg, \textit{sittable}) or nouns (\eg, \textit{armrest}), which describe how object parts could afford human daily actions and activities. We adopt certain common affordance categories from a comprehensive survey of visual affordance~\cite{hassanin2021visual}, \eg, \textit{containment}, \textit{sittable}, \textit{support}, \textit{openable}, \textit{rollable}, \textit{display}, and \textit{wrapgrasp}. 

However, the affordances defined in Hassanin \etal~\cite{hassanin2021visual} are coarse-grained--either at the object-level or scene-level. For example, ``\textit{openable}'' only indicates whether an object can be opened, but is unclear about which object part can \textit{afford} the object to be opened. To pursuit a fine-grained understanding of object affordance, we describe part-level affordance by manually constructing a one-to-multiple mapping from $479$ kinds of object part labels defined at the finest granularity in Mo \etal~\cite{mo2019partnet} to $24$ potential affordance labels. As a result, each affordance type--due to its cross-category nature--may be found on a variety of object part instances as illustrated in \cref{affordance_label}. For example, \textit{openable} is usually afforded by rotatable doors for unobstructed access. Under such criteria, the door frame of a dishwasher and the surface board of a door are both mapped to \textit{openable}. We also correct some fine-grained parts (\eg, different door handles (twistable, lever, \etc.)). Please refer to the \textit{appendix} for a full list of all affordance categories and mappings.

The PartNet dataset does not contain articulation information, making affordances such as \textit{openable} not geometrically distinguishable. Therefore, we generate a set of shapes with \textit{openable} affordance from the PartNet-Mobility dataset by capturing 3D shapes with various opening angles. 

\section{Experiments}

In this section, we design and conduct comprehensive experiments to evaluate the proposed method. \cref{fig:qualitative} visualizes our main results. We present the both quantitative and qualitative comparisons of baseline models and our model variants. To illustrate the generalizability of our approach, we evaluate the model generalization on novel objects. We further analyze the failure cases and propose potential improvement directions. Please refer to the \textit{appendix} for additional experimental results and analyses.

\subsection{Experimental Settings}

\paragraph{Benchmarks}

To benchmark \textit{PartAfford}, we curate a subset of samples from our constructed dataset. Specifically, we study the learning of most representative affordance categories ``sittable,'' ``support,'' and ``openable.'' As mentioned in \cref{sec:dataset}, a part can have multiple affordances. To ease the ambiguities in learning, we only choose the most prioritized affordance for each part. Below, we describe the statistics of objects related to these affordance categories; we discuss detailed reasons about why we benchmark these three affordance categories in \cref{sec:category}. 

``Sittable'': We collect all object instances that have the affordance label ``sittable''; most of them are chairs and sofas. Their part-level affordances belong to $\{sittable$, $backrest$, $armrest$, $framework\}$. We split the training, validation, and test set in the ratio of $7:1:2$. In total, we have $5,093$ instances for training and $1,457$ for test.

``Support'': Similarly, we collect 2 object categories: table and cabinet, whose part-level affordances belong to $\{support$, $framework\}$. There are $7,974$ instances for training and $2,279$ instances for test.  

``Openable'': We collect 4 object categories: refrigerator, dishwasher, washing machine, and microwave. Their affordances belong to $\{openable$, $framework$, $handle\}$. There are $807$ instances for training and $232$ instances for test.

\paragraph{Evaluation Metrics}

In \textit{PardAfford}, we evaluate the performances of part discovery (clustering similarity), 3D reconstruction, and affordance prediction. 

\begin{itemize}[leftmargin=*,noitemsep,nolistsep]
    \item \textbf{Part Discovery:} Prior work on 2D object discovery~\cite{locatello2020object,greff2019multi,burgess2019monet} employ Adjusted Rand Index (ARI) score to evaluate the clustering similarity between the reconstructed parts and ground-truth parts. However, compared with image ARI evaluation, the voxel value is binary. It makes voxel ARI intrinsically compare the number of voxels between parts, which cannot precisely reflect the part discovery accuracy. Thus, we use the Intersection over Union (IoU) to evaluate the part similarity. Specifically, we employ Hungarian matching to find the best matches between the reconstructed parts and ground truth parts using voxel IoU as the matching score. Then we compute the mean IoU by averaging the IoU between best matches.
    
    \item \textbf{3D Reconstruction:} We evaluate the overall 3D shape reconstruction quality using mean squared error (MSE).
    
    \item \textbf{Affordance Prediction:} Following~\cite{locatello2020object}, we use Average Precision (AP) to evaluate the affordance set prediction accuracy. A correct prediction means an exact match of the affordance label set.
\end{itemize}

\paragraph{Data Augmentation}

To increase the diversity and variety of the training data, we augment the data by randomly removing certain parts and their corresponding affordance labels on $30\%$ of training samples. We study the effects of data augmentation in ablation analysis and discuss the diversity of affordance composition in \cref{sec:diversity}.

\paragraph{Baselines and Ablations}

Since we are the first to propose and formulate \textit{PartAfford}, there is no previous work for us to make direct comparisons. Therefore, we compare with two designed baseline models and three variants of our method to evaluate the efficacy of the proposed method and its components:

\begin{itemize}[leftmargin=*,noitemsep,nolistsep]
    \item Slot MLP: a simple MLP-based baseline where we replace Slot Attention with an MLP that maps from the learned feature maps (resized and flattened) to the (now ordered) slot representation.
    
    \item IODINE: a baseline where we replace the Slot Attention with an object-centric learning method IODINE~\cite{greff2019multi} to abstract and cluster the encoded feature.
    \item Ours w/o Afford \& Cuboid \& Aug: our model variant only keeps the 3D part reconstruction branch and trains without data augmentation.
    \item Ours w/o Afford \& Aug: our model variant where we discard the affordance prediction branch and train without data augmentation. 
    \item Ours w/o Cuboid: our model variant that discards the cuboidal primitive regularization branch.
    \item Ours Full: our full model that trains with the data augmentation technique.
\end{itemize}

We keep the same hyperparameters for these models.

\subsection{Implementation Details}

\paragraph{Learning Strategy}

To stabilize the training, we split the training into two stages. In the first stage, we train the decoder only with 3D part reconstruction and affordance prediction branches. In the second stage, we add the cuboidal primitive regularization branch into joint training with a lower learning rate. 

\begin{figure}[t!]
	\centering
	\includegraphics[width=\linewidth]{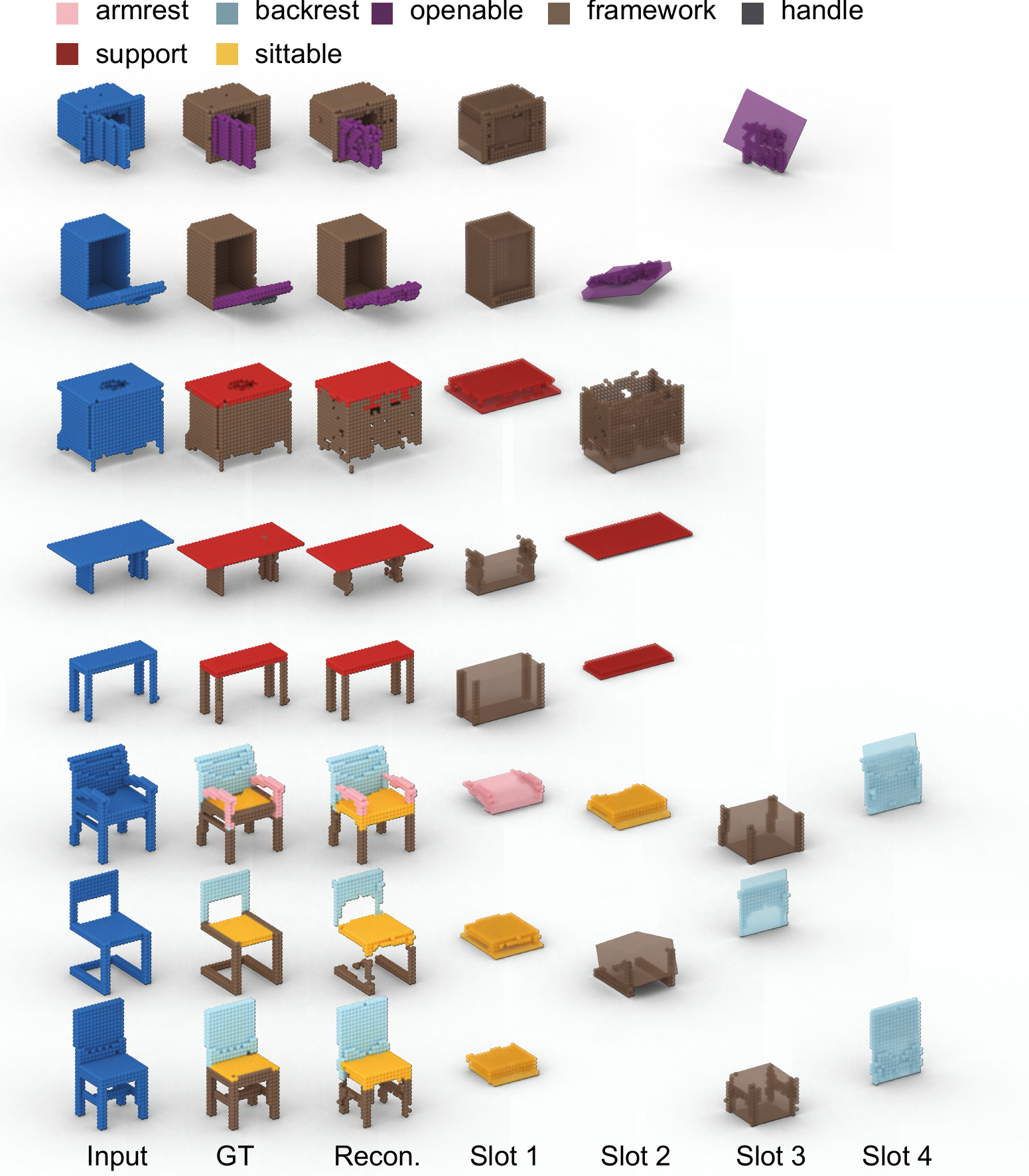}
	\caption{\textbf{Qualitative results of ``sittable,'' ``support,'' ``openable.''}}
	\label{fig:qualitative}
\end{figure}

\paragraph{Hyperparameter} We set learning rate as $4\times 10^{-4}$ for the first stage, $2\times 10^{-4}$ for the second stage, and apply Adam optimizer \cite{kingma2014adam} for optimization. It takes $5+9$ hours on $4$ RTX A6000 GPUs for two-stage full-model training of ``sittable''-related objects. We set the number of slots as $4$ for ``sittable,'' $3$ for ``openable,'' and $3$ for ``support''. For slot attention, we empirically set $T=3$. For the joint loss, we set $\lambda_{\text{recon}}=1.0, \lambda_{\text{pred}}=0.5, \lambda_{\text{cuboid}}=0.1$. We set the number of slots to the maximal number of affordance labels that appear in each affordance category. For example, we learn the ``sittable'' with $4$ slots, ``support'' with $2$ slots, and ``openable'' with $3$ slots. \cref{sec:slot} further discusses choices of the number of slots.

\begin{figure}[t!]
	\centering
	\begin{subfigure}[t]{\linewidth}
    	\includegraphics[width=\linewidth]{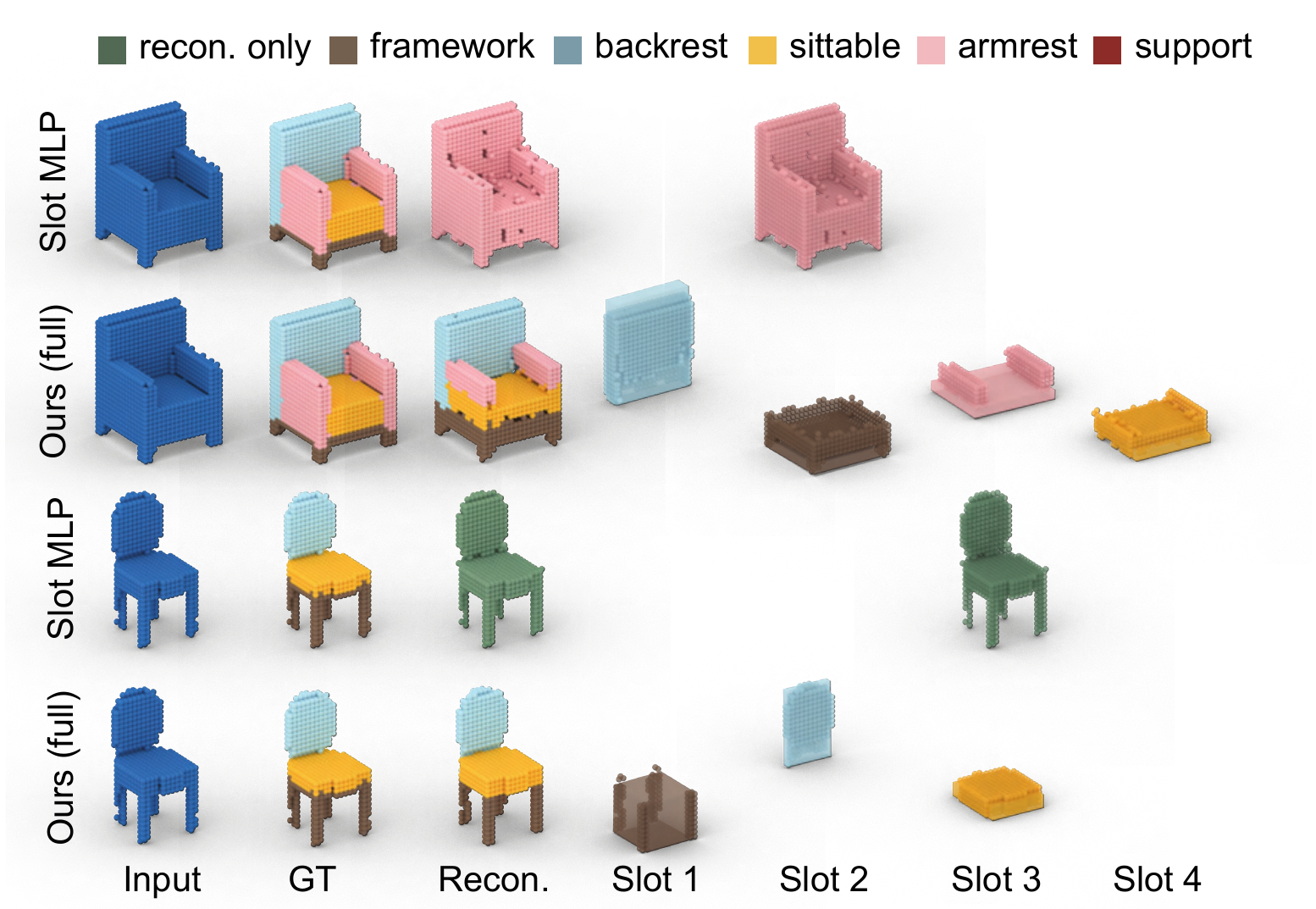}
    	\caption{Comparisons between the Slot MLP and our model.}
    	\label{fig:compare_slotmlp}
	\end{subfigure}
	\begin{subfigure}[t]{\linewidth}
    	\includegraphics[width=\linewidth]{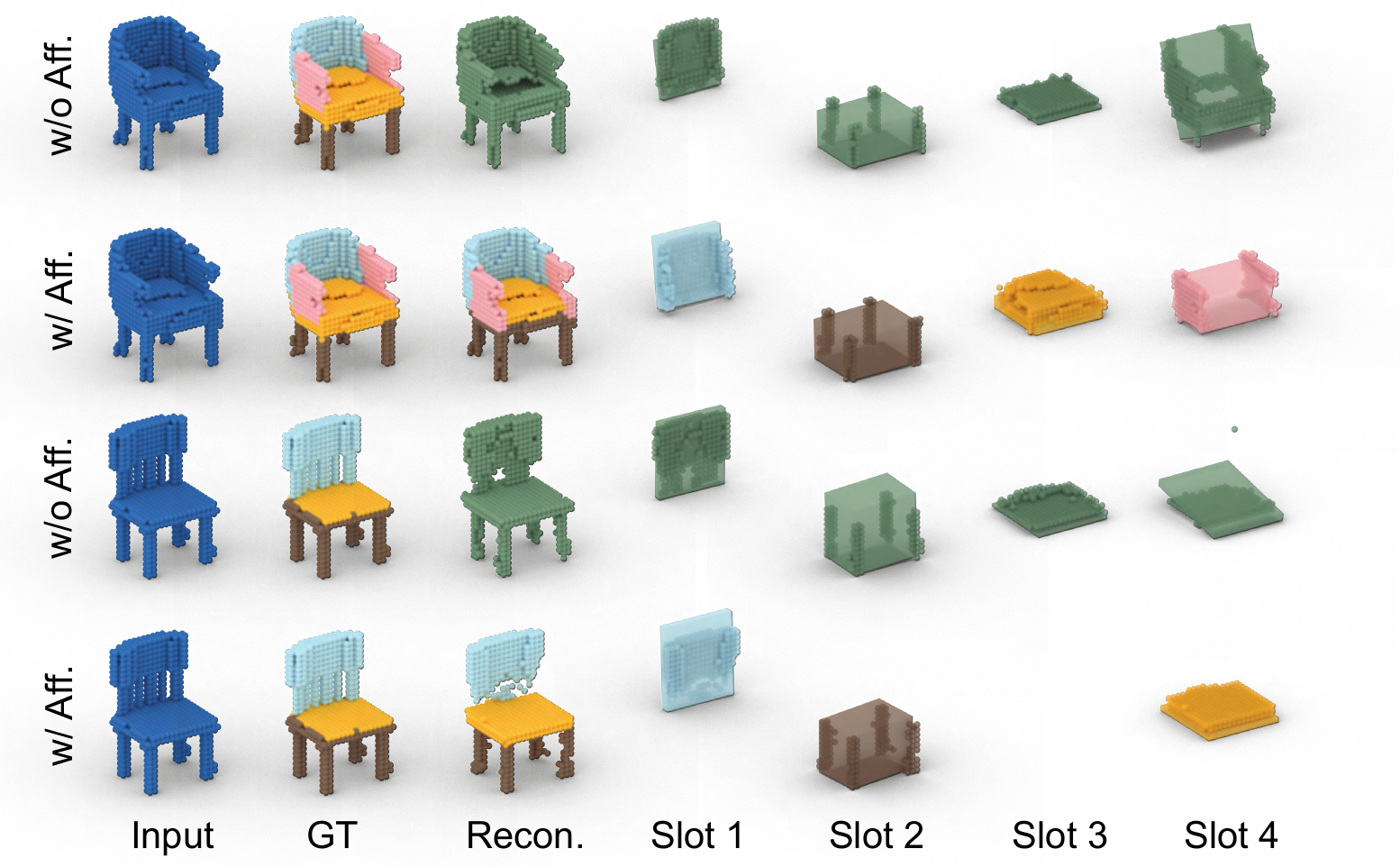}
    	\caption{Comparisons between the models without and with affordance prediction branch. The affordance prediction facilitates the abstraction of 3D shape (rows 1-2) and elimination of spare slots (rows 3-4).}
    	\label{fig:compare_afford}
	\end{subfigure}
	\begin{subfigure}[t]{\linewidth}
    	\includegraphics[width=\linewidth,trim={0 0 3.4cm 0}]{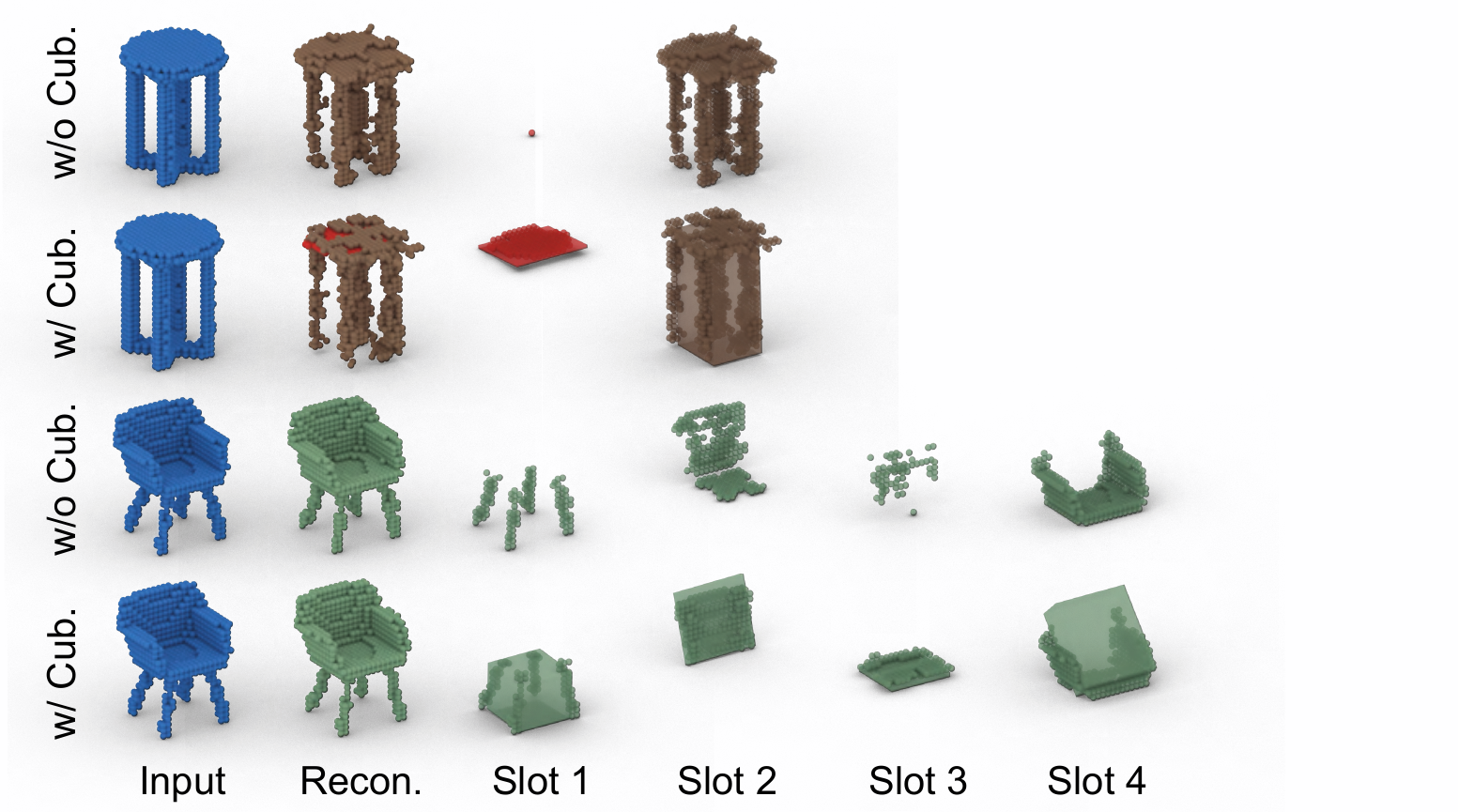}
    	\caption{Comparisons between the models without and with cuboidal primitive regularization branch. The cuboidal primitive regularization helps to guide and improve the segmentation.}
    	\label{fig:compare_cuboid}
	\end{subfigure}
	\caption{\textbf{Qualitative comparison results.}}
	\label{fig:qualitative_compare}
\end{figure}

\subsection{Quantitative Results}

\begin{table}[htbp]
    \centering
    \caption{Quantitative results on ``sittable''. We evaluate the mean IoU, mean squared error (MSE), and average precision (AP) on objects related to ``sittable.''}
    \label{tab:quantitative_sit}
    \resizebox{\hsize}{!}{%
    \begin{tabular}{c|c|c|c}
        \toprule
        \textbf{Model} &  mean IoU (\%) & MSE & AP (\%)\\
        \hline
        Slot MLP &21.5 & 0.0150 &\textbf{94.5}\\
        IODINE & 49.2 & 0.0102 & 92.5  \\
        Ours w/o Afford \& Cuboid \& Aug & 31.5 & 0.0112 & N/A \\
        Ours w/o Afford \& Aug & 39.4 & 0.0100 & N/A \\
        Ours w/o Cuboid & 55.3 & 0.0102 & 92.7 \\
        \hline
        Ours (full) & \textbf{57.3} & \textbf{0.0097} & 92.9  \\
        \bottomrule
    \end{tabular}
    }
\end{table}

As shown in \cref{tab:quantitative_sit,tab:quantitative_support}, we evaluate the quantitative performances of all the models. The proposed method outperforms the baselines by a large margin, which illustrates the outstanding abstraction capability of our approach. Note the Slot MLP achieves the best AP since it focuses only on set prediction but not part abstraction and discovery, as shown in \cref{fig:compare_slotmlp}. 

For ablations, affordance prediction and data augmentation significantly escalate the mean IoU since they help to improve the diversity of affordance composition, as also explained in \cref{sec:diversity}. Cuboidal primitive regularization branch also boosts the mean IoU, especially without affordance prediction and data augmentation, demonstrating that geometric priors play a crucial role in segmentation when data are not diverse enough. 

\begin{table}[htbp]
    \centering
    \caption{Quantitative results on ``support'' and ``openable.''} 
    \label{tab:quantitative_support}
    \resizebox{0.8\hsize}{!}{%
    \begin{tabular}{c|c|c|c|c}
        \toprule
        & \textbf{Model} &  mean IoU (\%) & MSE & AP (\%) \\
        \hline
        \multirow{2}*{support} & Ours w/o Cuboid & 51.3 & 0.0087 & \textbf{95.2} \\
        ~& Ours (full) & \textbf{52.7} & \textbf{0.0085} & 95.1  \\
        \hline
        \multirow{2}*{openable} & Ours w/o Cuboid & 46.7 & 0.0097 & 55.8 \\
        ~& Ours (full) & \textbf{47.6} & \textbf{0.0093} & \textbf{60.4}  \\
        \bottomrule
    \end{tabular}
    }
\end{table}

\subsection{Qualitative Results}

As shown in \cref{fig:qualitative}, our method is capable of segmenting the object into shapes and learning the corresponding affordances with sparse affordance set supervision. \cref{fig:qualitative_compare} compares different models. We summarize the main observations as: \textit{(i) the Slot MLP cannot segment the object input parts due to the lack of abstraction capability (\eg, \cref{fig:compare_slotmlp}); (ii) the affordance prediction facilitates the abstraction of 3D shape (\eg, rows 1-2 of \cref{fig:compare_afford}) and elimination of spare slots (\eg, rows 3-4 of \cref{fig:compare_afford}); (iii) the cuboidal primitive regularization helps to segment better primitives and avoid scattered voxels (\eg, \cref{fig:compare_cuboid}).} These observations validate the efficacy of various components in our model.


\subsection{Model Generalization}

\begin{figure}[t!]
	\centering
	\includegraphics[width=\linewidth,trim={1cm 0 0 0}]{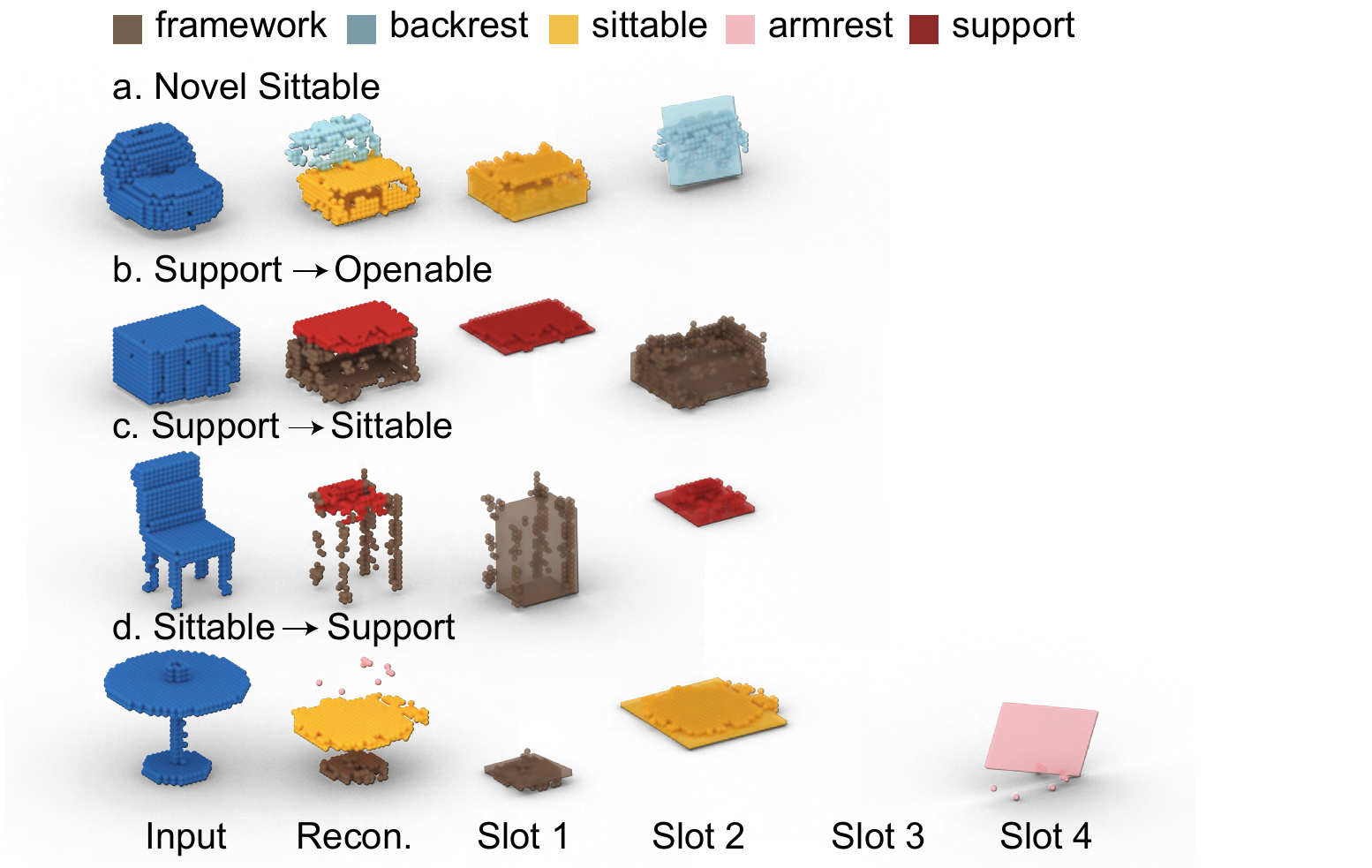}
	\caption{\textbf{Generalization results.} We show the results of testing the learned model on novel object shapes (a) and unseen categories (b-d). For example, (b) shows the result of learning with ``support'' and test on a ``openable'' object (\ie, a microwave). Although some reconstructions are not perfect, \textbf{the learned model can successfully identify the functional parts given a novel object.}}
	\label{fig:generalization}
\end{figure}

With the cross-category nature of affordance, we qualitatively test how the learned model can be generalized to novel objects and unseen categories. We conduct model generalization experiments by testing hard examples or objects from other categories. \cref{fig:generalization} shows some examples and demonstrates the learned model could be generalized to objects with diverse shapes. Although the reconstructions may not be perfect in part due to reconstruction bottleneck's impact on disentanglement quality ~\cite{engelcke2020reconstruction}, the learned model can successfully identify the functional parts given novel objects.

\subsection{Failure Cases}

\begin{figure}[t!]
	\centering
	\includegraphics[width=\linewidth, trim={0.5cm 0 0 0}]{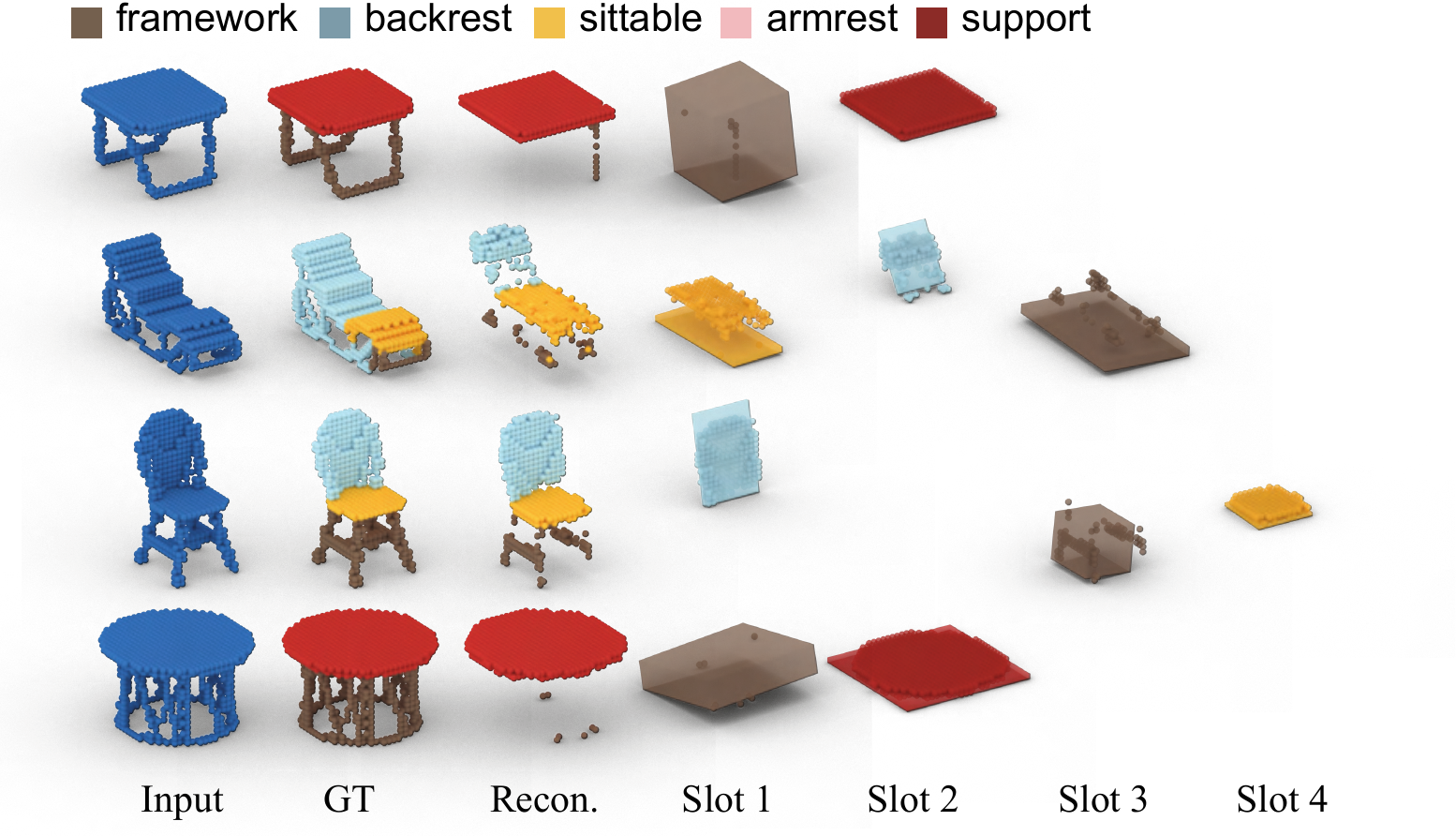}
	\caption{\textbf{Failure cases where the object are not well segmented or reconstructed.}}
	\label{fig:failure}
\end{figure}

We show some failure cases of our method in \cref{fig:failure}. For ``sittable'' and ``support,'' the failures are commonly caused by (i) the difficulties to reconstruct the fine-grained details of 3D objects with novel shapes; (ii) certain parts that violate the cuboid assumption thus hurts other components.

For objects in ``openable'' category, our model cannot discover and reconstruct ``handle,'' as shown in \cref{fig:qualitative}. This is because the objects with ``openable'' affordance come from various object categories with diverse shapes, making it challenging for the model to capture such complex mixtures of distributions and reconstruct the fine-grained 3D shapes, especially tiny parts (\ie, ``handle'').




\begin{figure}[t!]
	\centering
	\includegraphics[width=\linewidth, trim={0.7cm 0 0 0}]{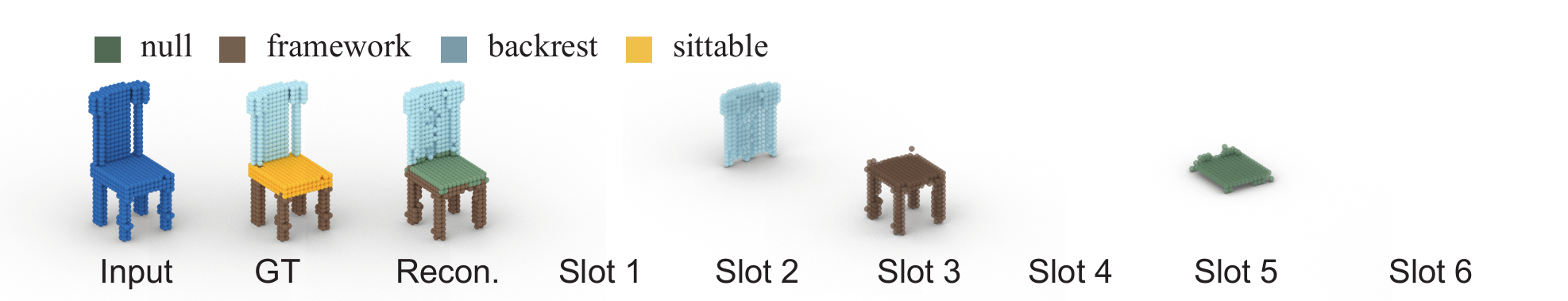}
	\caption{\textbf{Model performance when the number of slots increases.} The model learns the ``sittable'' in a ``null'' slot.}
	\label{fig:num_slot}
\end{figure}

\section{Further Analysis \& Discussion}

\subsection{Number of Slots}\label{sec:slot}

We set the slot number as the maximal number of affordance labels that appear in one category, which is different from previous object-centric learning algorithm~\cite{{locatello2020object}}, where the slot numbers could be arbitrary. This is because when we increase the number of slots, we increase the ambiguities during set matching at the same time. It prevents the model from learning accurate correspondence between affordance labels and parts. As shown in \cref{fig:num_slot}, the model learns the ``sittable'' in a ``null'' slot.

\subsection{Diversity of Affordance Composition and Learning Efficiency}\label{sec:diversity}

As shown in \cref{tab:quantitative_sit}, the performance of the proposed method greatly improves with data augmentation. The core reason is that the data augmentation increases the diversity of affordance composition on a large scale. For example, in experiments about ``sittable,'' most instances contain both ``sittable'' and ``backrest.'' It makes their affordance prediction interchangeable since the affordance set supervision does not provide contrasts for them. 

With data augmentation, the model can learn from contrasting samples where there only exist ``sittable.'' Therefore, it can learn better affordance correspondence with rich affordance composition, which provides enough contrasts. Similarly, humans could augment the learning samples by actively interacting with the environments, generating sufficient contrasting examples, and facilitating an efficient learning process. Therefore, we hope to imitate this superior capability into our model for efficient and active affordance learning in the future.

\subsection{Selection of Studied Affordances}\label{sec:category}

Although we annotate objects with 24 affordance categories, we study only three representative affordances in this work. The main reason is that we find it much more challenging to discover part-level affordance for some other affordance categories. For example, ``rollable'' and ``cutting'' usually connect to tiny object parts that are challenging to be segmented from objects, ``illumination'' and ``display'' cannot be distinguished from the objects without a deeper understanding of the visual appearance and reflections, ``pressable'' and ``pinchable'' require richer interactions to be discovered. In summary, it is either especially challenging to segment or requires more than geometric information (\eg, active interactions) to discover the other affordance categories. We leave the exploration of them for future work.

\subsection{Ambiguities in Affordance Learning}

Affordance is naturally ambiguous since the functions of object parts are rich. This work provides a well-defined benchmark to study how to learn affordance from accurate affordance definition and sparse set supervision. In another work, \cite{deng20213d} models the multiple affordances with a mixture of distributions. We believe our current learning framework could also be extended to learn multiple affordances by switching the one-hot affordance label to multi-hot labels.

\section{Conclusion}

We present \textit{PartAfford}, a new task in visual affordance research that aims at discovering part-level affordances from 3D shapes. We argue it provides a natural setting for learning visual affordance with sparse humanlike supervision. We propose a novel learning framework that discovers part-level affordances by leveraging only the affordance set supervision and geometric primitive regularization. With comprehensive experiments and analyses, we point out potential directions for incorporating visual appearance to facilitate better shape abstraction and combining with an active learning approach for efficient affordance learning.

{\small
\setstretch{0.96}
\bibliographystyle{ieee_fullname}
\bibliography{reference}
}

\clearpage
\appendix

\twocolumn[{%
\renewcommand\twocolumn[1][]{#1}%
\maketitle
\begin{center}
	\centering
    \captionsetup{type=figure}
	\includegraphics[width=\linewidth]{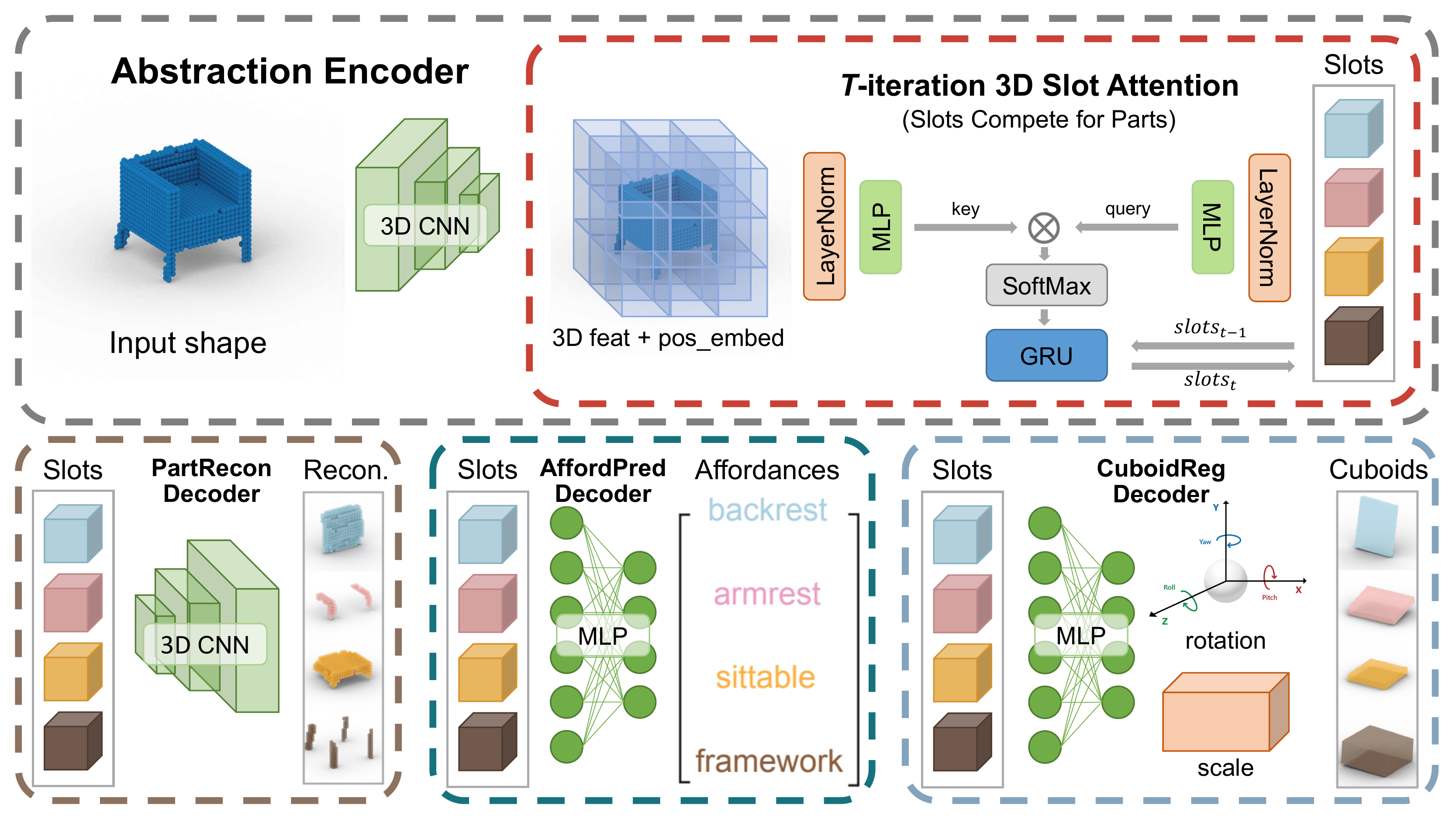}
	\captionof{figure}{Detailed diagram for sub-components of the proposed framework for PartAfford.}
	\label{fig:subcomponents}
\end{center}
\begin{center}
    \centering
    \captionsetup{type=figure}
    \includegraphics[width=\linewidth]{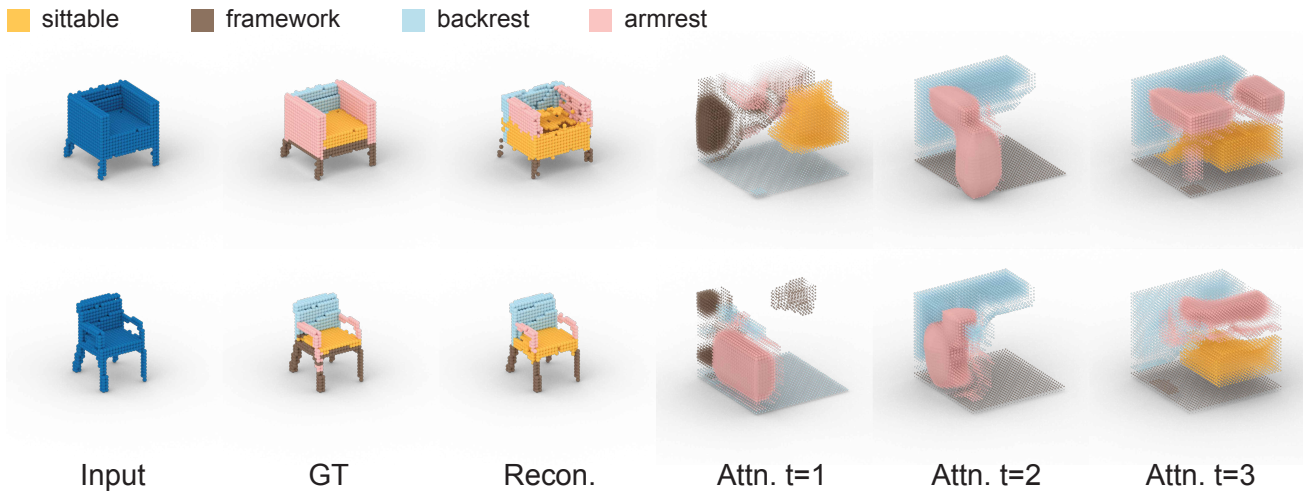}
    \captionof{figure}{Example affordance discovery model trained with $T=3$ attention iterations. Attention is visualized in various colors and point radius. Point is colored according to the predicted affordance label of the slot which attends the most to the point. Point radius positively depends on the maximal attention value across the slots. We use trilinear upsampling to rescale the attention mask to the input resolution ($32\times32\times32$).}
    \label{fig:attention}
\end{center}%
}]


\section{Attention Visualization}
Following the settings in \cite{locatello2020object}, we train the models using $T=3$ attention iterations in the slot attention module. In \cref{fig:attention}, we observe that each slot gradually attends to the correct part of the object as the number of attention iterations increases. The attention mask at $t=3$ depicts the silhouette of the reconstructed shape and parts.

\section{Ablation}
\subsection{Soft k-means}
Slot attention, as a generalized soft k-means algorithm, could be reduced to the soft k-means algorithm according to \cite{locatello2020object}. It turns out that the reduced model can achieve $23.0\%$ Mean IoU on "sittable" objects, which is slightly better than the Slot MLP baseline but significantly worse than the full model using slot attention ($57.3\%$). It reconstructs the whole shape with a quality (MSE: $0.0096$) on par with the full model (MSE: $0.0097$). Its affordance set prediction accuracy is the lowest among all models AP: $0.87$). Overall, the soft k-means algorithm cannot effectively segment the shape and discover affordance parts.

\subsection{Cuboid $L_1$ Norm}
For $m$-th slot, we predict its cuboid scale vector $\vs^m \in \R^3$. We regularize the cuboid loss by adding a cuboid scale penalty term. We adopt $L_1$ norm to compute the scale loss:
\begin{equation}
	\mathcal{L}_{\text{scale}} = \sum_m \| \vs^m\|_1.
\end{equation}
Without the cuboid $\mathcal{L}_1$ norm, some cuboids may not tightly wrap the reconstructed part if it is too slim.

\section{Additional Experimental Results}
\begin{figure}[h!]
    \centering
     \includegraphics[width=\linewidth]{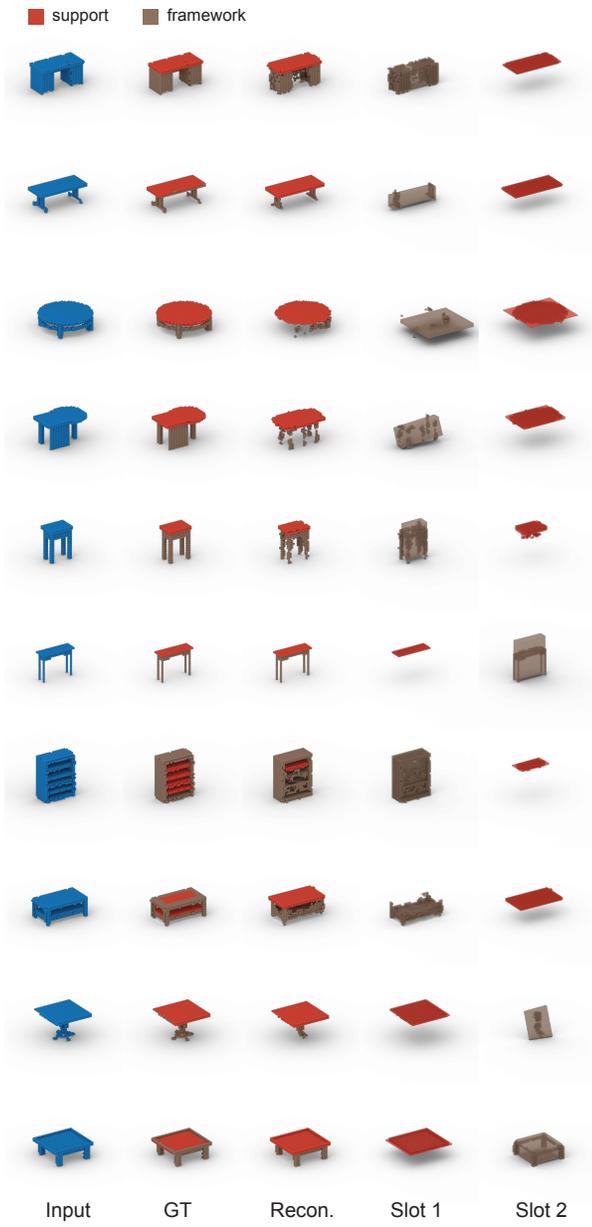}
    \caption{Additional qualitative results on "support" objects.}
    \label{fig:support}
\end{figure}

\begin{figure}[t!]
    \centering
    \includegraphics[width=\linewidth]{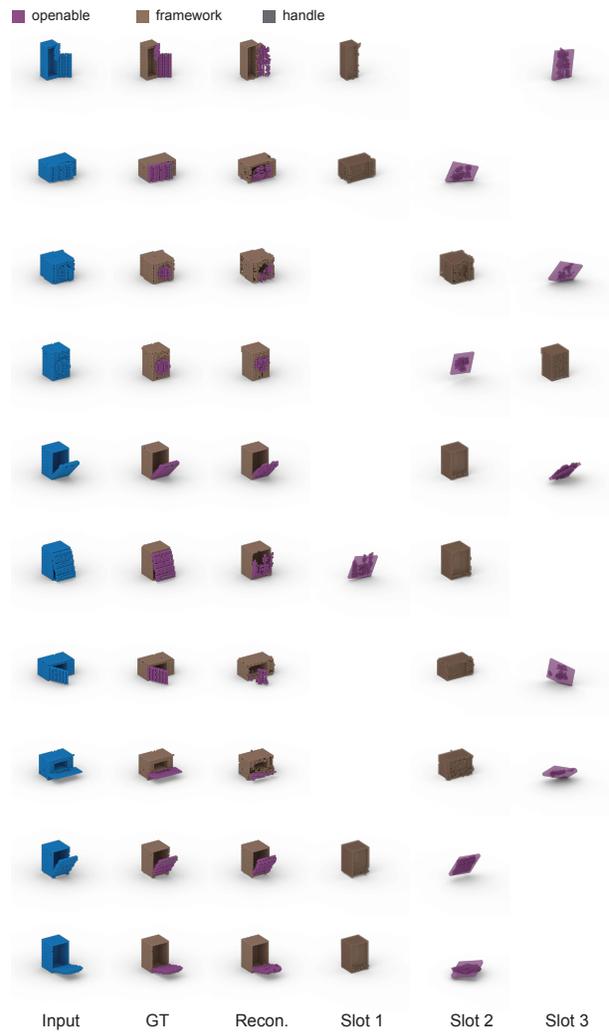}
    \caption{Additional qualitative results on "openable" objects.}
    \label{fig:openable}
\end{figure}

\begin{figure*}[t!]
    \centering
    \includegraphics[width=\linewidth]{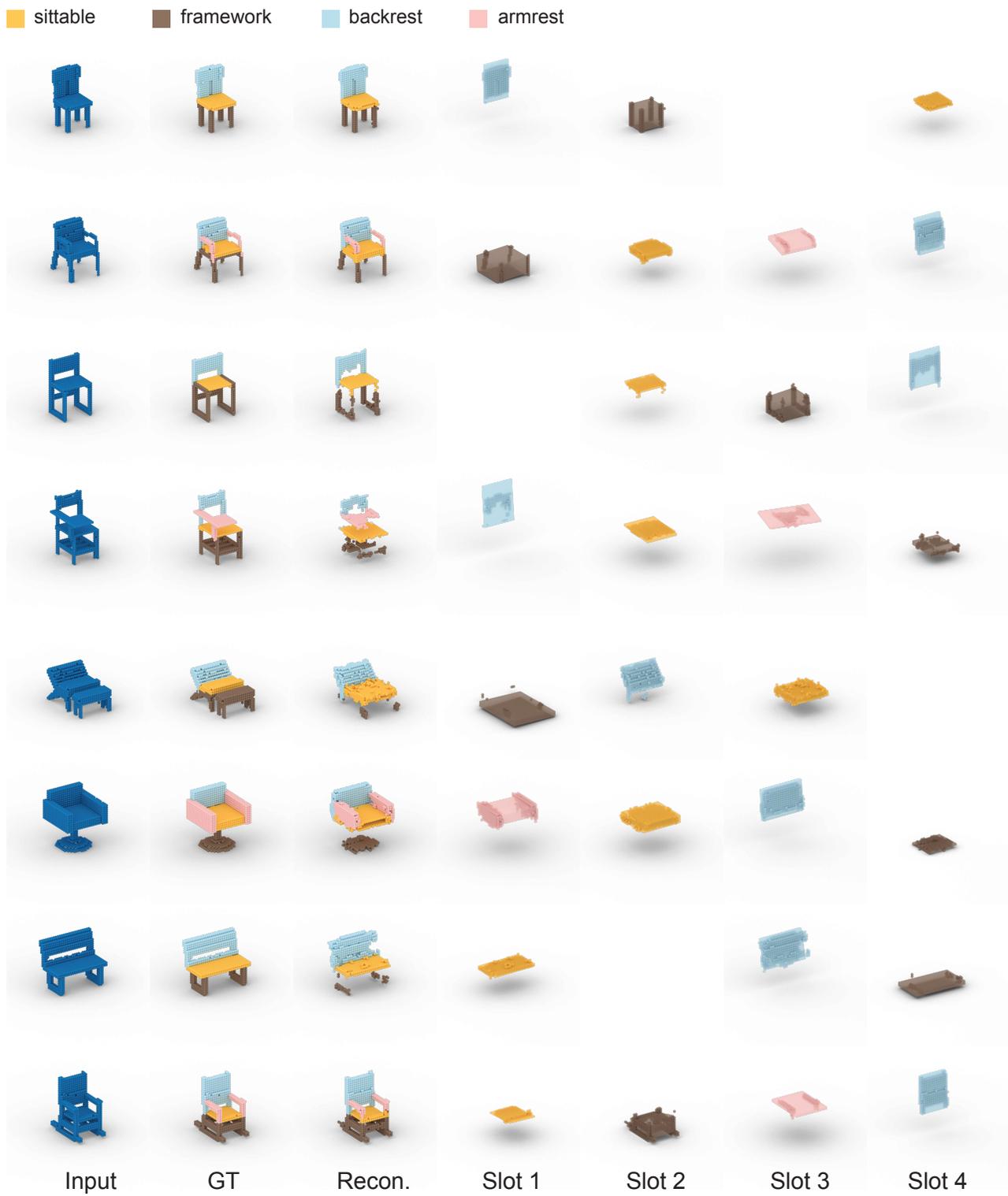}
    \caption{Additional qualitative results on "sittable" objects.}
    \label{fig:sittable}
\end{figure*}

We show additional qualitative results for our full model's affordance discovery results on "sittable" (\cref{fig:sittable}), "support" (\cref{fig:support}), "openable" (\cref{fig:openable}) objects, respectively.

\section{Part Affordance Dataset}

\subsection{Principles for Affordance Annotation}
To keep annotations consistent across object categories, we design a guideline for affordance annotation. Below are some general principles to annotate a leaf part of an object instance with affordances.

\textit{Multiple affordances.} A part can afford multiple kinds of human actions. For example, the seat of chair could afford \textit{sittable} for resting of human body or \textit{support} if one wants to place some books on it. We refer to ConceptNet~\cite{speer2017conceptnet}, a giant knowledge database, for annotating common usage of object parts.

\textit{Prioritized fine-grained affordances.} When there are multiple affordances labels for a part, we give the more fine-grained affordance label higher priority. For the chair seat in the example above, \textit{sittable} is prioritized compared with \textit{support} as it is a fine-grained support affordance for body resting.

\textit{Articulation-related affordances.} The PartNet dataset does not contain articulation information, which makes affordances such as \textit{openable} not geometrically distinguishable. Thus, we also generate a set of shapes with \textit{openable} affordance from the PartNet-Mobility dataset by capturing 3D shapes with various opening angles. More geometric variation helps models to learn articulation-related affordances. 

\subsection{Affordance Descriptions}
\label{affordance_description}
Our description for each affordance contains a brief
definition, some supplemental clarification and priority statement if needed, and some example leaf
nodes in the part hierarchy of various reasonable objects (full path from root to leaf).

\textit{sittable}: Indicates whether the object can be used for sitting. Anything sittable of course affords support, and the requirement for a supporting object to be sittable is that it must be both comfortable and safe for human seating. For example, a table is not sittable despite affording support because it is not comfortable. \textit{Sittable} is given priority over potential co-existing affordances like \textit{support}. 

	\Eg, chair$/$chair\_seat$/$seat\_surface.   
 
\textit{support}: A trait of objects which can safely keep other objects on top of themselves. Common characteristics of support-affording objects are that they are flat and can support multiple objects at once. A key distinction between support-affording objects and non-supporting objects is that support-affording objects will remain stable when other objects are placed on them. For example, a table is a supporting object because it is just as stable with objects on it as it is without. However, a stack of plates is not supporting because they become more unstable as you add more plates.

    \Eg, table$/$regular\_table$/$tabletop, \\  
    \indent ~~~~~~~ storage\_furniture$/$cabinet$/$shelf,\\
    \indent ~~~~~~~ bed$/$bed\_unit$/$bed\_sleep\_area. 
	
\textit{openable}: Parts which may be moved with a hinge-like mechanism on an articulated object. Openable objects do not need to afford handles, but usually handles can be found attached to the \textit{openable} part. \textit{Openable} parts are distinct from other moving parts in that they need to swing to some degree to be moved. \textit{Openable} is given priority over potential co-existing affordances like \textit{containment}. 
	
	\Eg, door$/$door\_body$/$surface\_board,\\  
    \indent ~~~~~~~ dish\_washer$/$body$/$door$/$door\_frame,\\  
    \indent ~~~~~~~ microwave$/$body$/$door.
    
\textit{backrest}: Objects which are reasonably designed for providing support to a person’s back. By reasonably designed, we mean either specifically (as in the back of a chair) or can afford back support if a person wanted to sit upright (like a headboard).

    \Eg, chair$/$chair\_back$/$back\_support.

\textit{armrest}: Objects which are specifically designed to support an arm. For example, a table can support a variety of things and is thus not an armrest. A chair’s arm is the perfect size for a human arm, so it must be an armrest.

    \Eg, chair$/$chair\_arm.

\textit{handle}: An object extension which affords the ability to open an attached ‘openable’ part. Handles are mostly grabbed with hand-wrapping, so it’s important to only afford ‘handle’ to parts which are specifically involved in an opening mechanism, like a door handle.

    \Eg, storage\_furniture$/$cabinet$/$cabinet\_door$/$handle, \\  
    \indent ~~~~~~~ table$/$regular\_table$/$table\_base$/$drawer\_base$/$\\  
    \indent ~~~~~~~~cabinet\_door$/$handle, \\  
    \indent ~~~~~~~ mug$/$handle,\\  
    \indent ~~~~~~~ dish\_washer$/$body$/$door$/$handle,\\  
    \indent ~~~~~~~ bag$/$bag\_handle.
    
\textit{framework}: Any object segment which either: a) helps to define the shape of the object as a whole or b) is an unaffording extension of the object or connector for other segments. For example, a hinge affords framework because it is integral to connects the door to the door frame. Overall, this affordance is the most general of all, so it should only be used when it clearly applies to either case of the definition. For example, most handles do not afford framework because it is both a small part of an object’s shape and already affords \textit{handle}. It has the least priority among potential co-existing affordances.

    \Eg, chair$/$chair\_base.

\textit{containment}: An affordance of object which can store physical items. The size of the physical items does not matter, so long as they are not too small. For example, anything that can only contain objects smaller than say a marble do not afford containment. Also, items that afford containment must afford security to the items they contain, such that they will not fall out.

    \Eg, table$/$regular\_table$/$table\_base$/$drawer\_base$/$drawer, \\  
    \indent ~~~~~~~ storage\_furniture$/$cabinet$/$drawer,\\  
    \indent ~~~~~~~ mug$/$container,\\  
    \indent ~~~~~~~ trash\_can$/$container,\\  
    \indent ~~~~~~~ refrigerator$/$body,\\  
    \indent ~~~~~~~ bowl$/$container, bag$/$bag\_body, \\
    \indent ~~~~~~~ bottle$/$normal\_bottle$/$body.

\textit{liquidcontainment}: A more specific version of the containment affordance. Objects that afford liquid-containment must be able to safely contain liquid. Examples of these are bottles, bath tubs, \etc.

    \Eg, mug$/$body,\\  
    \indent ~~~~~~~ bottle$/$normal\_bottle$/$body.

\textit{display}: Something which visualizes information for a useful purpose. Examples of these would be monitor screens or a clock surface.

    \Eg, display$/$display\_screen$/$screen,\\  
    \indent ~~~~~~~ laptop$/$screen\_side$/$screen, \\  
    \indent ~~~~~~~ clock$/$table\_clock$/$clock\_body$/$surface.

\textit{cutting}: The quality of being able to slice through other objects. Certain things that can cut are not considered to have \textit{cutting} affordance if it was used against its intended purpose, like smashing a glass vase. Cutting is only afforded to objects which are specifically designed for cutting, like a blade-edge.

    \Eg, cutting\_instrument$/$knife$/$blade\_side,\\  
    \indent ~~~~~~~ scissors$/$blade\_handle\_set$/$blade. 

\textit{pressable}: A mechanical feature of objects which either have buttons or can interact with a finger. Good examples of these are keyboard keys.

    \Eg, keyboard$/$key. 

\textit{hanging}: A part which can be hung on another object. These parts almost always only serve the purpose of hanging the rest of the entire object. An example of this would be a shoulder strap for a handbag.

    \Eg, bag$/$shoulder\_strap.

\textit{wrapgrasp}: The ‘wrap-grasp’ trait is afforded by parts which  are explicitly meant to be grabbed in a hand-wrapping motion. Just because a hand can wrap around an object part does not mean it affords wrap-grasp. It must be useful to grip the part in this way. An example of this would be a ladder rung, which a person is meant to wrap their hand around to climb the ladder.

    \Eg, cup,\\  
    \indent ~~~~~~~ bed$/$ladder$/$rung.

\textit{illumination}: The affordance of light emission. This only applies to object parts which are meant to light up a broad area. For example, a monitor screen does not afford illumination despite emitting light because it is not supposed to be used to light up the area around it.

	\Eg, lamp$/$table\_or\_floor\_lamp$/$lamp\_unit\\  
    \indent ~~~~~~~ $/$lamp\_head$/$light\_bulb.

\textit{lyable}: Indicates that a human can comfortably rest his$/$her entire body on the object. These objects are usually flat with a soft surface. \textit{Lyable} is given priority over potential co-existing affordances such as \textit{sittable} and \textit{support}. 

    \Eg, bed$/$bed\_unit$/$bed\_sleep\_area$/$mattress.

\textit{headrest}: An extension of an object which is oriented so a human head can rest comfortably on it. Examples of these are chair headrests or bedframe headrests.

    \Eg, bed$/$bed\_unit$/$bed\_sleep\_area\_pillow,\\  
    \indent ~~~~~~~ bed$/$bed\_unit$/$bed\_frame$/$headboard.

\textit{step}: A part which affords the human foot climbing or resting functionality. For example, a ladder rung is a step because it affords climbing with both hands and feet. A foot pedestal is also a step because it can be stood on or feet can be rested on it.

    \Eg, bed$/$ladder$/$rung.

\textit{pourable}: Meant for parts which liquid can flow out of. Things that are pourable may also be dependent on a mechanism for controlling flow, like a bottle cap or a knob.

    \Eg, bottle$/$normal\_bottle$/$mouth,\\  
    \indent ~~~~~~~ bottle$/$jug$/$body.

\textit{twistable}: These objects can either be detached or provide special functionality by twisting them in a clockwise or counterclockwise motion. Examples include bottle caps and knobs.

    \Eg, bottle$/$normal\_bottle$/$lid,\\  
    \indent ~~~~~~~ bottle$/$normal\_bottle$/$mouth.

\textit{rollable}: A part which can roll to move around. Exceptions to this affordance are objects which roll but stay fixed in place, like a rocking chair.

    \Eg, wheel.

\textit{lever}: Any handle which can rotate up to a point. For example, knobs rotate but are not levers because they do not provide handles. Levers must be treated differently from twistable objects or handles because if they are twisted too much they will break.

    \Eg, lever.

\textit{pinchable}: An object which is small enough such that it can be manipulated by pinching with two or more fingers. Things that are pinchable must not be heavy, and they usually fit inside the palm of a hand.

    \Eg, earbud.

\textit{audible}: Anything which emits sound. This does not include sound emitted indirectly, such as a door creaking when opened, which makes sound as a side-effect.

    \Eg, headphone$/$padding.
	


\subsection{License}
Our dataset is annotated based on PartNet (v0) \cite{mo2019partnet} and PartNet-Mobility (v2.0) \cite{xiang2020sapien}, both of which are licensed under the terms of the MIT License.

\clearpage

\newpage

\end{document}